\documentclass[runningheads]{llncs}
\usepackage[T1]{fontenc}
\usepackage{graphicx}
\usepackage[colorlinks=true,citecolor=blue,urlcolor=blue]{hyperref}
\usepackage{color}

\usepackage{amsmath}
\usepackage{amssymb}
\usepackage{booktabs}
\usepackage{ragged2e}

\begin{document}

\title{Generalizing Across Domains in Diabetic Retinopathy via Variational Autoencoders}

\titlerunning{VAE-DG}
%
\author{Sharon Chokuwa\thanks{Corresponding author} \and
Muhammad H. Khan}
%

\authorrunning{S. Chokuwa and M. Khan}

\institute{Mohamed Bin Zayed University of Artificial Intelligence, Abu Dhabi, UAE
\email{\{sharon.chokuwa,muhammad.haris\}@mbzuai.ac.ae}}

\maketitle              
\begin{abstract}
Domain generalization for Diabetic Retinopathy (DR) classification allows a model to adeptly classify retinal images from previously unseen domains with various imaging conditions and patient demographics, thereby enhancing its applicability in a wide range of clinical environments. In this study, we explore the inherent capacity of variational autoencoders to disentangle the latent space of fundus images, with an aim to obtain a more robust and adaptable domain-invariant representation that effectively tackles the domain shift encountered in DR datasets. Despite the simplicity of our approach, we explore the efficacy of this classical method and demonstrate its ability to outperform contemporary state-of-the-art approaches for this task using publicly available datasets. Our findings challenge the prevailing assumption that highly sophisticated methods for DR classification are inherently superior for domain generalization. This highlights the importance of considering simple methods and adapting them to the challenging task of  generalizing medical images, rather than solely relying on advanced techniques. 

\keywords{Domain Generalization  \and Diabetic Retinopathy \and Variational Autoencoder}
\end{abstract}

\section{Introduction}
Diabetic Retinopathy (DR) is a complication of Diabetes Mellitus (DM) which is characterized by impaired blood vessels in the eye due to elevated glucose levels, leading to swelling, leakage of blood and fluids, and potential ocular damage \cite{atwany2022deep}. With the global population infected with DM projected to reach approximately 700 million by 2045, DR is expected to persist as a prevalent complication of DM, particularly in the Middle East and North Africa as well as the Western Pacific regions \cite{teo2021global}. In general, the diagnosis of DR is based on the presence of four types of lesions, namely microaneurysms, hemorrhages, soft and hard exudates, and thus the categorization of DR typically comprises five classes, namely no DR, mild DR, moderate DR, severe DR, and proliferative DR. 

\begin{figure}[!h]
\centering
    \includegraphics[width=0.7\linewidth]{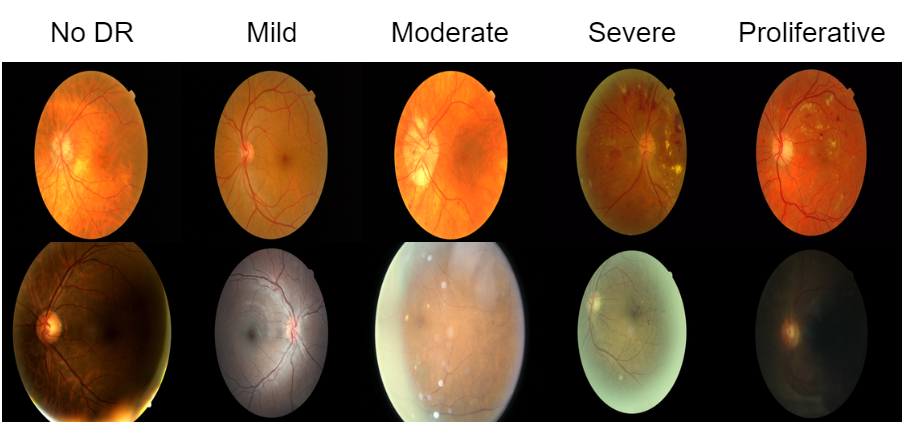}
    \caption{A sample of fundus images from MESSIDOR-2 (top row) and EyePACS (bottom row) datasets. For an untrained expert, it is challenging to sometimes visually see the differences between the different grades, making the DR classification task challenging. Each dataset exhibits a diverse range of variations in the presentation of fundus images and furthermore, the provided sample from the two domains clearly demonstrates a significant domain shift.}
    \label{fig:fundus_images}
\end{figure}

The conventional method of diagnosing DR relies on manual examination of retinal images by skilled ophthalmologists. However, this approach is known to involve time-intensive procedures, limited availability of trained professionals, and is susceptible to human error \cite{paisan2019deep,ting2016diabetic}. Deep learning methods have emerged as an effective solution for diagnosing DR, addressing the limitations associated with traditional approaches \cite{abramoff2020automated,ting2019artificial}. Despite the benefits offered by deep learning models, a major challenge they face is the issue of domain shift \cite{ting2019artificial}, which emanates from the oversimplified assumption of independence and identical distribution (i.i.d) between the training and testing data, leading to poor performance when these models are applied to new data from related but unseen distributions \cite{JMLR:v22:17-679,hendrycks2019benchmarking}. The variations in fundus image acquisition procedures and the diverse populations affected by DR result in a substantial domain shift as shown in Fig.~\ref{fig:fundus_images}, which greatly hinders the deployment of large-scale models since a slight variation of the data-generating process often foresees a drastic reduction in model performance  \cite{yang2021generalized}. 

Domain generalization (DG) is a line of research with the goal of handling the domain shift problem \cite{gulrajani2020search} under minimal assumptions. It only relies on multiple or seldom single source domain(s) to train a model that can generalize to data from unseen domains, whose distribution can be radically different from source domains.
To our knowledge, there exists a rather limited body of literature specifically addressing the problem of domain generalization for DR classification. Therefore, the investigation of DG for deep learning methods holds significant relevance in enhancing the accuracy of DR diagnosis across the various healthcare centers situated in different geographical locations.

In this paper, we propose our Variational Autoencoder for Domain Generalization (VAE-DG), which effectively manipulates the power of classical variational autoencoders (VAEs) \cite{Kingma_2019}, whose optimally disentangled latent space \cite{higgins2016beta} enables the model to generalize well to unseen domains in DR classification by effectively capturing essential shared information while selectively disregarding domain-specific variations. Through the acquisition of disentangled representations that separate domain-specific and domain-invariant features, VAE-DG significantly enhances the model's ability to generalize across different domains, leading to improved performance and robustness. Our main contributions in this work are as follows:
\begin{enumerate}
    \item We aim to inspire researchers to explore and leverage a wider spectrum of techniques, particularly simpler methods, in their pursuit of effective solutions for the challenging task of robustifying the DR classification problem.
    \item To our knowledge, we are the first to explore the potential of harnessing VAEs for learning cross-domain generalizable models for the Diabetic Retinopathy classification task. Our extensive analysis reveals compelling evidence of its superiority over the state-of-the-art techniques for the DG approaches in the DR classification task.
    \item We report our results using the training-domain validation criterion for model selection, which is an appropriate and widely-adopted model selection method for DG \cite{gulrajani2020search}, thereby rectifying the existing work's \cite{atwany2022DRGen} important limitations. To this end, we encourage future studies to conduct fair comparisons with our methodology, establishing a standard for evaluating advancements in DG for DR classification task.
\end{enumerate}

\section{Related Works}
\noindent\textbf{DG for DR classification:} DRGen \cite{atwany2022DRGen} could be considered as the first work that tackles the DG challenge in DR classification, by combining the Stochastic Weight Averaging Densely (SWAD) \cite{cha2021swad} and Fishr \cite{rame2022fishr} techniques. SWAD is a DG technique that promotes flatter minima and reduces gradient variance, while Fishr is a regularization method that aligns gradient variances across different source domains based on the relationship between gradient covariance, Hessian of the loss, and Fisher information. While the work by \cite{atwany2022DRGen} played a pivotal role in bringing attention to this problem task, it should be noted that the results presented by the authors were based on target-domain-validation, which does not align with the established protocols of evaluating DG methods, as outlined by the widely recognized DomainBed framework \cite{gulrajani2020search}. We rectify this limitation by adopting the appropriate model selection strategy of source-domain validation, in accordance with accepted practices in the field of DG research.

\noindent\textbf{DG using feature disentanglement:} DG approaches based on feature disentanglement aim to disentangle the feature representation into distinct components, including a domain-shared or invariant feature and a domain-specific feature \cite{wang2022generalizing}. Methods like \cite{ilse2020diva,nam2021reducing} focus on disentangling multiple factors of variation, such as domain information, category information, or style; while this can be beneficial for certain applications, this may lead to limited interpretability and difficulties in finding an optimal balance between the different disentangled factors causing complex training procedures. In contrast, our method provides a more holistic approach to feature disentanglement, and with appropriate regularization techniques, it can achieve stable training and straightforward optimization. \cite{peng2019domain,qiao2020learning,zhang2022towards} used fine-grained domain disentanglement, Unified Feature Disentanglement Network, and semantic-variational disentanglement, respectively, which introduces additional complexity to the model architecture, and often leads to increased computational costs during training and inference. On the contrary, our methodology which is both effective and simpler offers a more direct and efficient approach. 

\begin{figure}
\centering
    \includegraphics[width=0.7\linewidth]{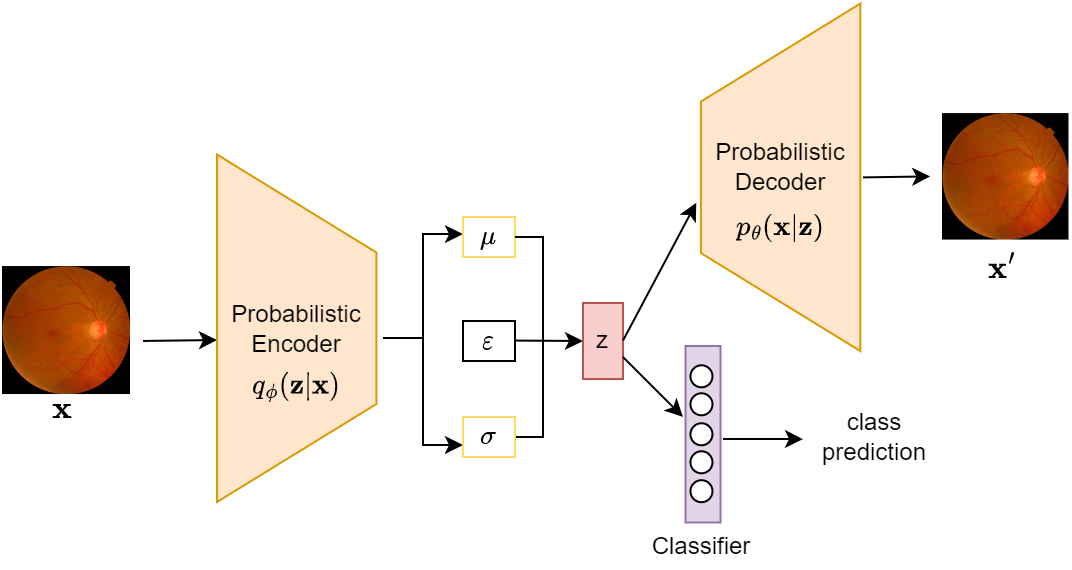}
    \caption{Overview of our proposed method VAE-DG for domain generalization with a variational autoencoder by manipulating the disentangled fundus image representations to achieve a domain generalization objective.}
    \label{fig:method}
\end{figure}

\section{Method}
\noindent\textbf{Overview:} In this section, we describe in detail on how we exploit conventional variational autoencoders to tackle the challenge of domain generalization by revisiting their operational principles and integrating them into our VAE-DG approach. This showcases their effectiveness in disentangling intricate DR datasets, within which we hypothesize that the optimally disentangled latent space contains domain-shared features, thereby yielding a substantial performance boost compared to existing domain generalization state-of-the-art methods. Our overall pipeline is shown in Fig.~\ref{fig:method}

\noindent\textbf{Problem settings:} Domain generalization for DR classification is defined within a framework that involves a collection of source domains denoted as $\{S_d\}_{d=1}^{N}$, where $N$ is the number of source domains. Each source domain $S_d = \{({x_i}^d, {y_i}^d)\}_{i=1}^{n}$ comprises i.i.d data points, sampled from a probability distribution $p(X_d, Y_d)$. $Y_d$ is the target random variable corresponding to the progression of DR, while $X_d$ is the input fundus image random variable, with each data point $({x_i}^d, {y_i}^d)$ representing an observation from its respective domain. The primary objective in domain generalization thus becomes acquiring a predictor that exhibits robust performance on an unseen target domain $T_d$ \cite{gulrajani2020search}.

\noindent\textbf{Proposed method (VAE-DG):} To achieve domain generalization using our VAE-DG, we manipulate two variables (from the pooled source domains $\{S_d\}_{d=1}^{N}$) which are the input fundus image $X_d$ and the latent variable $Z_d$. When we consider only singular data points, $z_i$ is drawn from the distribution  $z_i \sim p(z)$ and  $x_i$ is drawn from $x_i\sim p(x|z)$, and their joint distribution is given by $p(x,z)=p(x|z)p(z)$. The main goal of this probabilistic model becomes an inference problem of learning a distribution $p(z|x)$ of some latent variables from which we can then sample to generate new fundus images which we will denote as $x^\prime$. We know that this posterior distribution $p(z|x)$ can be obtained using Bayes Theorem \cite{joyce2003bayes}.

However, we utilize a 256-dimensional fundus latent vector whose marginal $p({x})$ requires exponential computational time and hence becomes intractable, therefore, instead of directly calculating $p_\theta({z}|{x})$, we resort to Variational Inference \cite{blei2017variational} such that we approximate this posterior with a tractable distribution $q_\phi({z}|{x})$ which has a functional form. We use the Gaussian distribution as the approximation such that the problem decomposes to learning the parameters $\phi$ = $(\mu,\sigma^2)$ instead of $\theta$. By incorporating this Gaussian prior as a constraint on the learned latent variables, our VAE-DG is coerced into disentangling the underlying factors of variation in the data. We can then use Kullback-Leibler (KL) divergence, to measure how well the approximation is close to the true distribution. By minimizing the KL divergence, we simultaneously approximate $p_\theta({z}|{x})$ and the manipulation of the KL divergence expression (the complete derivation of which is beyond the scope of this discussion but can be found in \cite{odaibo2019tutorial}), we obtain Equation~\ref{eq:4}:
    \begin{equation}
        \label{eq:4}
        \log p_\theta(x) - D_{\text{KL}} \left(q_{\phi}(z|x) || p_\theta(x)\right) = \mathbb{E}_z \left[{\log p{_\theta}(x|z)}\right] - D_{\text{KL}} \left(q_{\phi}(z|x) || p(z) \right)
    \end{equation}
where; $\mathbb{E}_z \left[{\log p{_\theta}(x|z)}\right] - D_{\text{KL}} \left(q_{\phi}(z|x) || p_\theta(z) \right)$ is known as the Evidence Lower Bound (ELBO), the former term thus becomes the lower bound on the log evidence. Subsequently, if we maximize the ELBO we thus indirectly minimize $D_{\text{KL}} \left(q_{\phi}(z|x) || p_\theta(x)\right)$. Therefore, the objective function of a classical variational autoencoder can be expressed as: 
    \begin{equation}
        \label{eq:5}
            \mathcal{L}(\theta, \phi; x) = -\mathbb{E}_{q_{\phi}(z|x)} \left[\log p_{\theta}(x|z)\right] + D_{\text{KL}} \left(q_{\phi}(z|x) || p(z)\right) 
    \end{equation}
where the objective function is with respect to $\theta$ and $\phi$ which are the learnable parameters of the generative and inference models, respectively \cite{kingma2013auto,Kingma_2019}. 

For our VAE-DG we couple the classical variational autoencoder objective $\mathcal{L}(\theta, \phi; x)$ with empirical risk minimization $\sum_{i=1}^{n} \ell(f(x_i), y_i)$ \cite{vapnik1999overview} to ensure the optimization of the original target task as illustrated in Equation \ref{eq:9}, while simultaneously manipulating the domain-invariant latent variables acquired from the probabilistic encoder. Our final objective function consists of three distinct terms; the first term, denoted by $-\mathbb{E}_{q_{\phi}(z|x)} \left[\log p_{\theta}(x|z)\right]$, serves as the reconstruction term, which quantifies the difference between the original fundus image $x_i$ and the reconstructed ${x_i}^\prime$. The second term, $\beta D_{\text{KL}} \left(q_{\phi}(z|x) || p(z)\right)$, is the regularizer term that minimizes the KL divergence between the encoder distribution $q_{\phi}(z|x)$ and the prior distribution $p(z)$, thereby promoting the learned latent representation ${z_i}$ to follow the prior distribution. The strength of this regularization is controlled by the hyperparameter $\beta$. The third term, $\sum_{i=1}^{n} \ell(f(x_i), y_i)$, assesses the difference between the true class labels $y_i$ and the predicted class labels $f(x_i)$, subsequently, the parameter $\alpha$ serves as a weight for this term. 
\begin{equation}
\label{eq:9}
\mathcal{L} = -\mathbb{E}_{q_{\phi}(z|x)} \left[\log p_{\theta}(x|z)\right] + \beta D_{\text{KL}} \left(q_{\phi}(z|x) || p(z)\right) - \alpha\sum_{i=1}^{n} \ell(f(x_i), y_i)
\end{equation}
To optimize $\mathcal{L}$ we use stochastic gradient descent with an incorporation of the alternate optimization trick \cite{Kingma_2019} since we need to learn the parameters for both $\theta$ and $\phi$.

\subsection{Experiments}
\noindent\textbf{Datasets:} We utilized four openly accessible datasets, namely EyePACS \cite{EyePACS}, APTOS \cite{APTOS}, Messidor \cite{MESSIDOR}, and Messidor-2 \cite{MESSIDOR} which according to their sources were obtained from different locations and populations, resulting in a notable domain shift due to variations in instruments, conditions, settings, and environmental contexts across datasets. Each dataset comprises of five distinct classes with the exception of Messidor, which lacks class 5 images. The dataset distribution for these sources is 88702, 3657, 1200, and 1744, respectively. The original images vary in size but are standardized to 224x224 pixels. Due to the inherent characteristics of real-world datasets, there exists an imbalance in class representation across all datasets with class 0 being the most dominant and class 4 the rarest.

\noindent\textbf{Implementation and evaluation criteria:}
Our choice for the encoder architecture involves the Imagenet pretrained ResNet-50 \cite{he2016deep} as the backbone. This is substantiated by existing literature \cite{matsoukas2022makes}, wherein the employment of transfer learning, despite the domain gap, has been demonstrated to accelerate the process of developing effective models even in medical imaging. We jointly trained on three source domains, with 0.2 of the source domains as the validation set, and finally evaluate on the unseen target domain using the best training-domain-validation model, this way we truly evaluate the domain generalizability of our model. The model is trained for 15,000 steps, with Adam optimizer, a learning rate of 0.0001,  256 dimensional $z$ latent vector, and a batch size of 66 from the three source domains. To combat class imbalance we utilize resampling. $\beta$ and $\alpha$ are set as 50,000 to achieve a similar weighting with the magnitude of the reconstruction term. Accuracy is used as the evaluation metric in line with the established DG benchmarks \cite{gulrajani2020search}. All our experiments were run on  24GB Quadro RTX 6000 GPU. Our code is available at \url{https://github.com/sharonchokuwa/VAE-DG}.

\noindent\textbf{Baselines:}
We compare our method with the naive Empirical Risk Minimization (ERM) \cite{gulrajani2020search,vapnik1999overview} and with state-of-the-art domain generalization methods for this problem task mainly DRGen \cite{atwany2022DRGen} and Fishr \cite{rame2022fishr}. To ensure a fair comparison, we adopt the same backbone and learning rate for all methods, except for DRGen; where we reproduce it using the original proposed learning rate of 0.0005, as the performance decreased when using 0.0001. The other method-specific hyperparameters were kept constant as proposed in the respective works. 
\begin{table}[ht]
\caption{Comparison between our proposed method with domain generalization methods for DR classification. Each experiment was repeated thrice, employing distinct random seeds (0, 1, 2), and the average accuracy (Avg.) and corresponding standard deviation are reported for each target domain.}
\footnotesize
\setlength{\tabcolsep}{3pt}
\label{tab:comparison}
\centering
\begin{tabular}{llllll}
\toprule
Method & Aptos & EyePACS & Messidor & Messidor-2 & Avg. \\
\midrule
\raggedright \vspace{0.2mm} ERM  & $63.75 \pm 5.5$ & $70.22 \pm 1.6$ & $\textbf{66.11} \pm 0.8$ & $67.38 \pm 1.0$ & $66.86 \pm 2.2$ \\
\raggedright \vspace{0.2mm} DRGen  & $57.06 \pm 0.9$ & $72.52 \pm 1.3$ & $61.25 \pm 4.2$ & $49.16 \pm 16.3$ & $60.00 \pm 5.7$ \\
\raggedright \vspace{0.2mm} Fishr  & $62.89 \pm 5.0$ & $71.92 \pm 1.3$ & $65.69 \pm 1.1$ & $63.54\pm 3.8$ & $66.01 \pm 2.8$ \\
\raggedright \vspace{0.2mm} VAE-DG  & $\textbf{66.14}\pm 1.1$  & $\textbf{72.74}\pm 1.0$  & $65.90 \pm 0.7$  & $\textbf{67.67} \pm 2.0$  & 
$\textbf{68.11} \pm \textbf{1.2}$ \\
\midrule
\multicolumn{2}{l}{\textit{Oracle Results}} \\
\raggedright \vspace{0.2mm} VAE-DG  & $68.54\pm 2.5$  & $74.30\pm 0.2$  & $66.39 \pm 1.3$  & $70.27 \pm 1.2$  & $69.87 \pm 1.3$ \\
\bottomrule
\end{tabular}
\label{tab:main}
\end{table}

\noindent\textbf{Results and Discussion:} Table~\ref{tab:main} indicates that VAE-DG exhibits the highest average accuracy of $68.11\pm1.2$\%, which represents an 8.11\% improvement over DRGen, 2.1\% over Fishr, and 1.3\% over ERM. Furthermore, VAE-DG demonstrates superior performance across most domains (APTOS, EyePACS, and Messidor-2) and exhibits the lowest standard error of 1.2\%, indicating its relative robustness compared to the other methods. VAE-DG's enhanced performance solidifies the advantageous characteristics of this simpler approach whose latent space facilitates the explicit disentangling of domain-specific and domain-invariant features, ultimately improving target domain generalization. The oracle results \cite{gulrajani2020search} of VAE-DG are presented as a reference for the upper bound of the method, rather than for direct comparison, indicating that our proposed method achieves a 1.8\% reduction compared to the upper bound.

ERM outperforms more sophisticated methods (DRGen and Fishr) because it is a simple approach and does not make strong assumptions about source-target domain relationships; it focuses on optimizing performance on available source domains and leveraging multiple domains to capture a wider range of variations, showcasing its ability to generalize to unseen target domains (if the domain shift is small \cite{gulrajani2020search}).

Overall, the relatively poor performances of DRGen and Fishr methods which attain 60.00\% and 66.01\% average accuracies respectively can be attributed to the fact that these methods often impose specific constraints or assumptions about the domain shift, which could limit their performance in scenarios that deviate from those assumptions. The lack of robustness of such methods with variations in the data is also vindicated by the large standard error (16.3\%) for DRGen's Messidor-2 domain performance. 

In contrast to the findings of \cite{atwany2022DRGen}, our extended analysis presented in Table~\ref{tab:extended_analysis} reveals a significant decline in model performance by 23.14\% when incorporating SWAD, aligning with \cite{cha2021swad}'s observation that SWAD is not a perfect or theoretically guaranteed solver for flat minima. We explored the influence of a larger network architecture (ResNet-152) and the obtained results indicate that a larger network architecture can improve image reconstruction quality but has a negative impact on the primary DG objective, as evidenced by the 1.5\% drop.

\begin{table}[ht]
\caption{Analysis and ablation studies. Average accuracy (Avg.) values represent the mean accuracy obtained from three independent trials. The "Diff." column indicates the performance variation compared to our main experiments shown in Table~\ref{tab:main}. A decrease in performance is denoted by ($\downarrow$), while an increase is denoted by ($\uparrow$). 
}
\footnotesize
\setlength{\tabcolsep}{1.5pt}
\label{tab:extended_analysis}
\centering
\begin{tabular}{llcccccc}
\toprule
& APTOS & EyePACS & Messidor & Messidor-2 & Avg. & Diff. \\
\midrule
\multicolumn{2}{l}{\textbf{Extended Analysis}} \\
\midrule
VAE-DG ResNet-152 & 61.45$\pm$8.2 & 71.44$\pm$3.1 & 65.94$\pm$1.0 & 67.81$\pm$2.6 & 66.66$\pm$3.7 & 1.45($\downarrow$) \\
\midrule
VAE-DG + SWAD & 55.66$\pm$8.8 & 73.52$\pm$0.0 & 34.24$\pm$12.2 & 16.48$\pm$12.0 & 44.97$\pm$8.3 & 23.14($\downarrow$) \\
ERM + SWAD & 54.93$\pm$0.6 & 71.35$\pm$0.5 & 64.76$\pm$0.7 & 58.48$\pm$3.1 & 62.38$\pm$1.2 & 4.5($\downarrow$)\\
\midrule
\multicolumn{2}{l}{\textbf{Ablations Studies}} \\
\midrule
Latent-dim 64 & 62.15$\pm$3.1 & 73.80$\pm$0.4 & 66.42$\pm$2.1 & 68.98$\pm$3.0 & 67.84$\pm$2.2 & 0.27($\downarrow$) \\
Latent-dim 128 & 62.61$\pm$3.5 & 73.64$\pm$0.6 & 66.60$\pm$1.9 & 66.09$\pm$2.2 & 67.23$\pm$2.0 & 0.88($\downarrow$) \\
\midrule
Fixed latent space & 63.87$\pm$0.6 & 73.44$\pm$0.8 & 66.46$\pm$0.6 & 69.39$\pm$0.8 & 68.29$\pm$0.7 & 0.18($\uparrow$) \\
\midrule
$\beta$, $\alpha$ = 10,000 & 64.38$\pm$1.8 & 73.17$\pm$0.5 & 65.42$\pm$0.4 & 69.27$\pm$4.0 & 68.06$\pm$1.7 & 0.05($\downarrow$) \\
$\beta$, $\alpha$ = 100,000 & 62.50$\pm$3.5 & 72.30$\pm$1.6 & 66.56$\pm$1.3 & 67.88$\pm$1.0 & 67.31$\pm$1.8 & 0.80($\downarrow$) \\
\midrule
No Recon Loss & 63.44$\pm$3.9 & 70.62$\pm$0.8 & 66.25$\pm$0.8 & 65.21$\pm$1.4 & 66.38$\pm$1.7 & 1.73($\downarrow$) \\
No KL Divergence & 68.29$\pm$2.3 & 69.98$\pm$4.3 & 66.60$\pm$1.1 & 66.93$\pm$1.6 & 67.95$\pm$2.3 & 0.17($\downarrow$) \\
\bottomrule
\end{tabular}
\end{table}

\noindent\textbf{Ablation studies:}
In order to comprehensively assess the individual contributions of each component towards our DG objective, we conducted ablation studies, as summarized in Table~\ref{tab:extended_analysis}. Our investigation encompassed the following aspects: (i) Latent-Dim: varying the size of the latent dimension [64, 128, 256], (ii) Fixed latent space: evaluating the impact of a fixed latent dimension, (ii) determining the impact of the weighting for the KL divergence and classification terms ($\beta$ and $\alpha$), (iii) assessing the effect of the reconstruction term, and (iv) examining the influence of the KL divergence term.

We noticed that a larger latent dimension of 256 leads to higher results, potentially due to its ability to effectively bottleneck information while preserving essential features. The performance difference between a fixed latent vector and a randomly sampled one is not very large, although using a fixed latent space reduces the standard error by nearly half, suggesting that randomly sampled vectors introduce additional variability that hinders the disentanglement of domain-invariant features. Notably, removing the reconstruction and KL divergence terms in the model's objective leads to a decrease in performance, emphasizing the importance of incorporating these regularizations. Furthermore, experimentation with $\beta$ and $\alpha$ values within the range of [10,000, 50,000, 100,000] reveals that excessively high or low values are suboptimal.

\section{Conclusion}
In this paper, we explored the potential of classical variational autoencoders for domain generalization in Diabetic Retinopathy classification. We demonstrate that this simple approach provides effective results and outperforms contemporary state-of-the-art methods. By strictly following the established evaluation protocols of DG, we also addressed the important limitations in the evaluations of the existing method. Our study encourages the medical imaging community to consider simpler methods in order to realize robust models.

\bibliographystyle{splncs04}
\bibliography{ref.bib}

\end{document}